\def\year{2022}\relax
\documentclass[letterpaper]{article} % DO NOT CHANGE THIS
\usepackage{aaai22}  % DO NOT CHANGE THIS
\usepackage{times}  % DO NOT CHANGE THIS
\usepackage{helvet}  % DO NOT CHANGE THIS
\usepackage{courier}  % DO NOT CHANGE THIS
\usepackage[hyphens]{url}  % DO NOT CHANGE THIS
\usepackage{graphicx} % DO NOT CHANGE THIS
\usepackage{natbib}  % DO NOT CHANGE THIS AND DO NOT ADD ANY OPTIONS TO IT
\usepackage{caption} % DO NOT CHANGE THIS AND DO NOT ADD ANY OPTIONS TO IT
\DeclareCaptionStyle{ruled}{labelfont=normalfont,labelsep=colon,strut=off} % DO NOT CHANGE THIS
\usepackage{algorithm}
\usepackage{algorithmic}
\usepackage{adjustbox}
\usepackage{colortbl}
\usepackage{multirow}
\usepackage{xcolor}
\usepackage{amsmath}
\usepackage{amssymb}
\usepackage{enumitem}

\definecolor{highlight}{cmyk}{0,0,0,0.29}

\usepackage{newfloat}
\usepackage{listings}
\floatstyle{ruled}
\newfloat{listing}{tb}{lst}{}
\floatname{listing}{Listing}

\title{Cross-Region Domain Adaptation for Class-level Alignment}
\author {
    Zhijie Wang\textsuperscript{\rm 1},
    Xing Liu\textsuperscript{\rm 1},
    Masanori Suganuma\textsuperscript{\rm 1,2},
    Takayuki Okatani\textsuperscript{\rm 1,2}
}
\affiliations {
    \textsuperscript{\rm 1} Graduate School of Information Sciences, Tohoku University \\
    \textsuperscript{\rm 2} RIKEN Center for AIP \\
    \{zhijie, ryu, suganuma, okatani\}@vision.is.tohoku.ac.jp
}
\usepackage{bibentry}
\begin{document}

\maketitle

%%%%%%%%% ABSTRACT
\begin{abstract}
% labor intensiveなSSのアノテーションを避けるため，別ドメインのデータ間（CGで作ったデータから実画像）でのDAする方法が盛んに研究されてきている．native dataで訓練した場合との性能のギャップはまだ大きく，改善の余地は大きい．本研究では，adv.trainingとself-trainingを融合する方法を提案．target domainの
Semantic segmentation requires a lot of training data, which necessitates costly annotation. There have been many studies on unsupervised domain adaptation (UDA) from one domain to another, e.g., from computer graphics to real images. However, there is still a gap in accuracy between UDA and supervised training on native domain data. It is arguably attributable to class-level misalignment between the source and target domain data. 
To cope with this, we propose a method that applies adversarial training to {\em align two feature distributions in the target domain}. It uses a self-training framework to split the image into two regions (i.e., trusted and untrusted), which form two distributions to align in the feature space.
% it uses a self-training framework to specify the feature distributions to align as those from different image regions. 
We term this approach {\em cross-region adaptation} (CRA) to distinguish from the previous methods of aligning different domain distributions, which we call {\em cross-domain adaptation} (CDA). CRA can be applied after any CDA method. Experimental results show that this always improves the accuracy of the combined CDA method, having updated the state-of-the-art. 
% Experimental results show that the method outperforms the current state-of-the-art in the widely used benchmark tests, i.e., GTA5, SYNTHIA, or Synscapes $\rightarrow$ Cityscapes.
% Compared to classic algorithms, methods based on convolutional neural networks (CNNs) have made significant progress in semantic segmentation tasks assisted by massive data with manual pixel-wise annotations, which is time-consuming and labor-intensive. Some virtual environments like video games can quickly generate images with annotations. Still, models trained with these synthetic data often perform poorly in real scenarios, which is known as domain shift problem. Self-training is an effective way to ease this problem, which selects regions with trusted predictions as the pseudo labels for training. However, existing methods rarely pay attention to those regions with untrusted predictions and these regions are usually discarded during the self-training, which could hurt the final performance. To better utilize the trusted regions, we propose a entropy-based region splitting method to distinguish trusted regions and and untrusted regions and a cross-region adaptation (CRA) training method that can align the feature distribution between them. Extensive experiments on several baseline models and benchmark tasks show that our cross-region adaptation training can outperform other approaches with state-of-the-art performance.
\end{abstract}
        
%%%%%%%%% BODY TEXT
\section{Introduction}

Semantic image segmentation is one of the fundamental problems of computer vision \cite{minaee2021image}. 
% It is used as a component for various applications like scene understanding, medical image analysis, autonomous driving, and so on. 
Methods employing
% convolutional neural network (CNNs) 
neural networks have achieved great success for the problem, which presume the availability of a large amount of labeled data. As manual annotation is costly, researchers have considered using synthetic images generated by computer graphics, for which precise pixel-level annotation is readily available \cite{gta5,synthia,synscapes}.

However, it is generally hard to apply 
% CNNs 
neural networks trained with synthetic data to real images because of the distributional difference between synthetic and real images. Many studies have been conducted on domain adaptation \cite{long2015learning,sun2016deep} to cope with the difference known as domain shift. Among several problem settings, the one that attracts researchers' most interest is unsupervised domain adaptation (UDA) \cite{Saito_2018_CVPR,Baktashmotlagh_2013_ICCV}. 
It is to train a model using labeled data in a domain (called the source domain) so that it will work well on data in a different domain (called a target domain) for which labels are not available. 

There are currently two approaches to UDA for semantic segmentation. One is {\em adversarial training} \cite{adaptseg,advent,dada,fada}, which attempts to obtain domain-invariant features by aligning the data distributions of the source and target domains. An issue with this approach is that while it may be easy to align the two distributions as a whole, it is hard to attain class-level alignment, leading to suboptimal results. The other approach is {\em self-training}, in which a teacher model trained with the labeled data in the source domain is used to generate pseudo labels of the target domain data and use them for training a student model \cite{zou2018unsupervised,zou2019confidence,li2019bidirectional,zhang2019category}. An issue with this approach is that pseudo labels could be inaccurate, which will lead to unsatisfactory performance.

A promising direction for further improvements is to {\em integrate} adversarial training and self-training, as is attempted by recent studies \cite{intrada, fada}. This paper proposes a new approach in the same direction, which applies adversarial training to align two feature distributions {\em in} the target domain. Using a self-training framework, it splits target domain images into two regions, thereby specifying the feature distributions to align. 

% Generally, adversarial training can be used to align any two distributions; it 
% merely 
% trains a feature extractor to extract indistinguishable features from them. The proposed method aligns the feature distributions of {\em untrusted} region and {\em trusted} region in target domain images. These two regions are selected by thresholding the confidence of the class prediction made by an initial model at each image pixel. 

\begin{table}[t]
\caption{The proposed method (cross-region adaptation, or CRA) can be employed after any existing method (which we call cross-domain adaptation, or CDA), always leading to performance improvement.
% our CRA based on different baseline methods. 
UDA from GTA5 $\rightarrow$ Cityscapes.}
\label{tab:summary}
% \begin{center}
% \begin{adjustbox}{width=0.8\columnwidth}
\centering
\small
\begin{tabular}{l|c|c|c}
\hline
Method & Base CDA & +CRA & $\Delta$ \\
\hline
AdaptSegNet~\cite{adaptseg} & 42.4 & 43.4 & +1.0 \\
ADVENT~\cite{advent} & 43.8 & 46.7 & +2.9 \\ 
FADA~\cite{fada} & 50.1 & 52.2 & +2.1 \\ 
IAST~\cite{iast} & 52.2 & 54.1 & +1.9 \\ 
ProDA~\cite{proda} & 57.5 & 58.6 & +1.1 \\
\hline
\end{tabular}
\label{tab:introduction}
\end{table}

\begin{figure*}[t]
\centering
\includegraphics[width=0.9\linewidth]{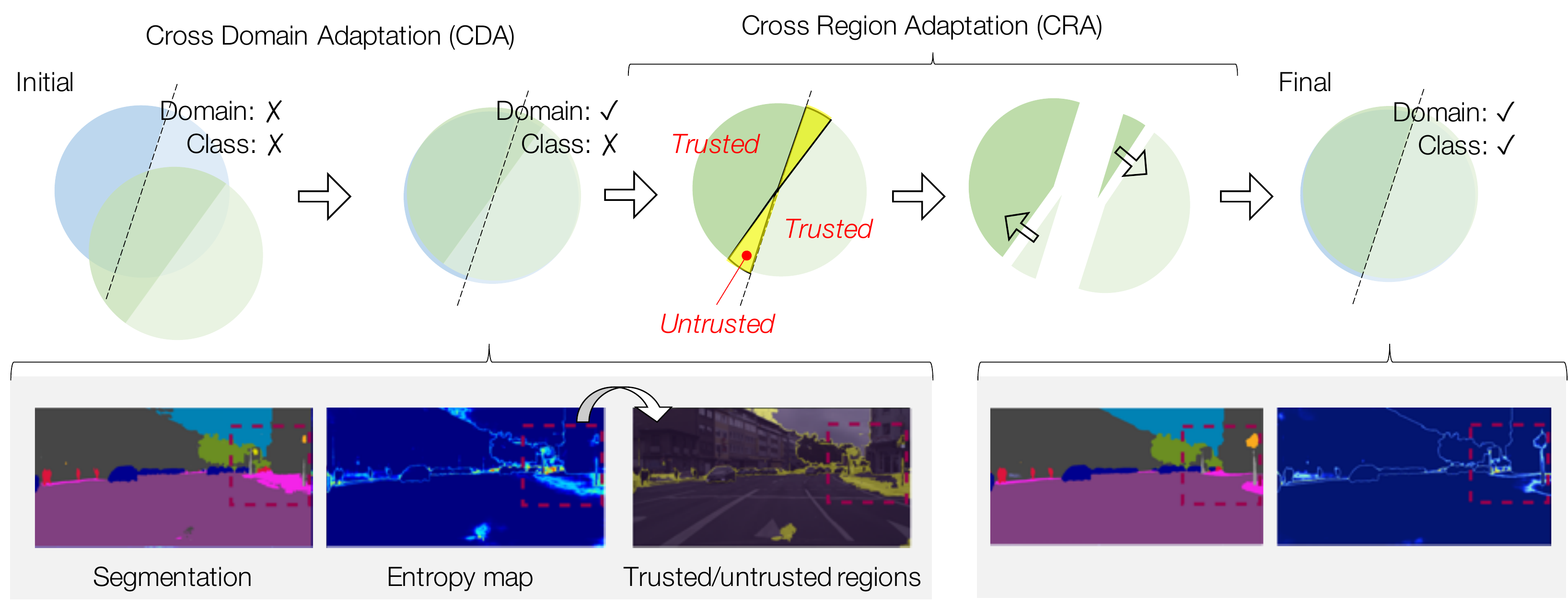} %pics/intro.png}
\caption{Illustration of our method. 
% For the left and right columns, from top to bottom are segmentation maps, entropy maps and feature distributions. For the middle column, from top to bottom are original target image, split target image and the feature distribution. 
The cross-domain adaptation aligns the two feature distributions of different domains by adversarial training, but it may not align their classes. The cross-region adaptation splits the image into trusted and untrusted regions based on the confidence map and aligns the feature distributions of the two regions by adversarial training.}
% entropy-based region splitting method to find the trusted regions and untrusted regions and align the feature distribution of them via proposed cross-region adaptation training. } 
\label{pic:intro}
\end{figure*}

We term this approach {\em cross-region adaptation} (CRA) to distinguish from the conventional method of aligning different domain distributions, which we will refer to as {\em cross-domain adaptation} (CDA). Its objective is to resolve class-level misalignment between the source and target domain data; see Fig.~\ref{pic:intro}. CRA can be employed after any existing CDA method, which we will show always improves the accuracy of the baseline CDA; a summary is shown in Table \ref{tab:summary}.

We show through experiments the effectiveness of the proposed approach on three benchmark tasks, GTA5 $\rightarrow$ Cityscapes, SYNTHIA $\rightarrow$ Cityscapes, and Synscapes $\rightarrow$ Cityscapes. The results show that our CRA, employed upon any existing baseline CDA, improves its performance. This is also the case with the CDA method achieving the current state-of-the-art \cite{proda}, meaning we have updated it.

\section{Related Work}

% Our work is mainly related to semantic segmentation, unsupervised domain adaptation, and entropy-based uncertainty estimation. 

%-------------------------------------------------------------------------
\subsection{Semantic Segmentation}

Methods based on convolutional neural networks (CNNs) have been the most successful for semantic segmentation. FCN is the first fully convolutional network for the pixel-level classification task proposed in a pioneering work \cite{fcn}. Later, UNet \cite{unet} and SegNet \cite{segnet} were proposed, which are the networks consisting of an encoder and a decoder, leading to better performance. Architectural designs have been extensively studied since then such as DeepLab \cite{deeplab}, PSPNet \cite{pspnet}, UperNet \cite{upernet}, DANet \cite{danet}, and EncNet \cite{encnet}, to name a few.
% improve the accuracy by mining the relationship between pixels through the attention mechanism.

%-------------------------------------------------------------------------
\subsection{Unsupervised Domain Adaptation}

There are two approaches to unsupervised domain adaptation (UDA) for semantic segmentation, i.e., adversarial training and self-training. The former mainly attempts to decrease a domain gap by performing adversarial training in feature space~\cite{chen2019learning,fada}, in input space~\cite{gong2019dlow,lee2018diverse}, or in output space~\cite{adaptseg}. The core idea of self-training is to generate  pseudo labels for target domain samples and use them for training the model~\cite{PFAN_2019_CVPR}. CBST~\cite{zou2018unsupervised} and CRST~\cite{zou2019confidence} conduct class-balanced self-training and confidence-regularized self-training, respectively, to generate better pseudo labels.
    
Recently, several attempts have been made to combine adversarial training and self-training to improve the performance \cite{li2019bidirectional,zheng2020unsupervised,iast}. BLF~\cite{li2019bidirectional} uses pseudo labels without any filtering, which could lead to label errors. AdaptMR~\cite{zheng2020unsupervised} filters pseudo labels but ignores the filtered-out pixels, which could retain valuable information. IAST~\cite{iast} uses filtered pseudo-labels for supervised training and applied entropy minimization, a semi-supervised learning method, to the filtered-out pixels. Our method splits the target-domain images into trusted and untrusted regions based on an entropy-based confidence map and applies adversarial training to these two regions' features, aiming to achieve finer alignment of the source- and target-domain distributions. 
Thus, our method is orthogonal to the above methods and experimental results show that it performs better.

% However, these methods always involve tedious and time-consuming multi-rounds training for better performance, and also lack effective use of regions are not marked as pseudo label.

%-------------------------------------------------------------------------
\subsection{Use of Predictive Uncertainty for Segmentation}

Many methods utilize the entropy of predicted class probabilities to measure the uncertainty of the model's prediction for better training.
% CEAL~\cite{wang2016cost} uses entropy as a metric to select high-confidence samples from the unlabeled set for feature learning of the classification task. RTN~\cite{NIPS2016_ac627ab1} exploits the principle of entropy minimization for refining classifier adaptation. {\color{red} \bf [How are they related to our problem? There should be more studies to cite, if we consider such general studies.]} 
% In UDA for semantic segmentation, 
ADVENT~\cite{advent} proposes to use the entropy for adversarial training and also for unsupervised training on the target domain data (i.e., entropy-minimization).
% generates {\color{blue} weighted self-information maps} from the entropy of the class probabilities and uses them for adversarial training or direct entropy minimization. 
% DADA uses a branch to estimate the depth information, then, it fuses the depth and the entropy as a new self-information map and conducts adversarial training on this new self-information map.
% ???
DADA \cite{dada} estimates scene depth from the same input images at training time. It aligns the source and target distribution in the standard feature space and jointly in the depth space, aiming for more accurate alignment.
% {\color{blue} DADA~\cite{dada} attempts  adversarial adaptation on depth-aware maps, which are combinations of the weighted entropy maps and the depth predictions.} 
ESL~\cite{saporta2020esl} uses the entropy to assess the confidence of the prediction better and filter pseudo labels for self-training while ignoring unselected pixels. IntraDA~\cite{intrada} splits target domain data into easy and hard samples based on the entropy and performs intra-domain adversarial training. Although our method is similar in using adversarial training within the target domain, ours consider aligning features from different image regions; more importantly, its performance is much higher. 

%-------------------------------------------------------------------------
\section{Proposed Method}

% We propose a method that integrates adversarial training and self-training to dissolve class-level misalignment of the source and target domain features. {\color{blue} \em Maybe this sentence can be removed.}

%-------------------------------------------------------------------------
% \subsection{Adversarial Domain Adaptation}
\subsection{Revisiting Adversarial Domain Adaptation}

% \subsubsection{Cross Domain Adaptation: Basic Approach}
Before explaining our method, we revisit the standard adversarial training for cross-domain adaptation (CDA). 
The problem is stated as follows. We are given labeled data $X_s = \{(x^{(s)}, y^{(s)})\}$ of the source domain and unlabeled data $X_t =\{x^{(t)}\}$ of the target domain. We assume here the two domains share the same $K$ semantic classes to predict. We wish to train a segmentation network $G=C\circ F$, where $C$ is a classifier and $F$ is a feature extractor. We first train $G$ on $X_s$ by minimizing 
the cross-entropy loss:
\begin{equation}\label{eq:seg_loss}
\mathcal{L}_{seg}^{cda} = -\sum_{i=1}^{H \times W}\sum_{k=1}^{K}y_{ik}^{(s)}\log p_{ik}^{(s)},
\end{equation}
where $p_{ik}^{(s)}=G(x^{(s)})$ is the softmax probability of pixel $i$ belonging to class $k$ and $y_{ik}^{(s)}$ is the ground-truth one-hot label. 
To make $G$ works well also with the target domain images, we consider a discriminator $D$ that distinguishes the domain (i.e., source or target) from an input feature $f=F(x)$.
Freezing $C$, we then train $F$ and $D$ in an adversarial fashion, aiming at aligning the distributions of the two domains in the feature space.

% Our solution is to first use entropy as a criterion to find pixel features that may be misclassified and to label them as the untrusted regions. This is followed by further adversarial training to the target data so that the untrusted pixel feature distribution will be well aligned with the trusted one and result better performance.
% \\
% {\color{blav} [Xing] The proposed training strategy has two steps, which are i) cross domain adaptation and ii) cross region adaptation. We describe them in the following sections. [Okatani] Yes, I'm aware of it. Let me handle it.}

\subsection{Cross Region Adaptation (CRA)}
% \textcolor{blue}{Why Cross Region Adaptation Help?}}

If CDA is successful, the source and target distributions should be well aligned. However, there is no guarantee that this achieves the class-level alignment between the two domains, as is well recognized in the community \cite{fada}. We consider reducing this class-level misalignment after CDA. 

\subsubsection{Outline of the Method.}

Specifically, suppose we have conducted CDA using an existing UDA method,
% \textcolor{blue}{finished the cross domain adaptation (CDA) training,}
% applied adversarial training to the given data, 
obtaining the segmentation network $G$. If class-level alignment is inaccurate, errors should occur around the class boundaries in the feature space, as shown in Fig.~\ref{pic:intro}. We wish to reduce the number of wrongly classified pixels, or equivalently, to move the pixels currently on the wrong side of the class boundary to the right side in the feature space. 

We may be able to identify 
% where such errors emerge in the images 
such erroneous pixels
using the uncertainty of the class prediction. We employ here the entropy of the predicted class probability for this purpose. To be specific, we first classify each image pixel into two classes, {\em trusted} and {\em untrusted}, by thresholding the entropy. The details will be explained in Sec.~\ref{ch:method}. 

% To move the pixels currently on the wrong side of the class boundary to the right side in the feature space, 
We then align the feature distributions of the trusted and untrusted pixels. To do so, we conduct adversarial training to these distributions within the target domain. 
% It is similar to the standard adversarial training, i.e., CDA; 
While CDA, the standard UDA method, aligns the feature distributions of source- and target-domain images, CRA aligns the distributions of the trusted and untrusted pixels in the target domain.
% CRA treats the feature distribution for the trusted pixels as the source domain distribution and that for the untrusted pixels as the target domain distribution. 
In parallel to this adversarial training, we employ the standard self-training in which we train the network with the pseudo labels on the trusted pixels provided by its earlier model.

% {\color{blue} We also use pseudo labels together...}

% {\color{blue} Note that our method performs the standard self-training using only trusted pixels with $\mathcal{L}_{seg}^{cra}$ alternately with the alignment of trusted and untrusted pixels with $\mathcal{L}_{adv}^{cra}$ and $\mathcal{L}_{D}^{cra}$.}

\subsubsection{Underlying Assumptions (Why Should CRA Work?)}

The above procedure should achieve our goal of reducing the number of wrongly classified pixels, if the following two conditions are met:
\begin{itemize} \em
% [noitemsep,topsep=5pt,wide=\parindent]
    \item All the trusted pixels are correctly classified, and some of the untrusted pixels are not.
    \item The population of the untrusted pixels is relatively small, as shown in Fig.3.
\end{itemize}
If the first condition is met, CRA attempts to align the untrusted pixels, some of which may be wrongly classified, to the trusted (thus correctly classified) pixels in the feature space, achieving the above goal. In practice, the mere alignment of the two could lead to a worse result; for instance, it could reshape the class boundary, leading to more misclassifications. This is not the case if the second condition is met. As there are a small number of untrusted pixels, their alignment to the trusted ones will not significantly impact the class boundary, which is supported by many trusted pixels. We will experimentally show the approach's effectiveness, thereby showing that the above conditions are satisfied in practice.

% The boundary determined by the source data is not good for the target data, which is a recently well-recognized problem in the community. 

% Although the above method contributes to performance improvement, the class-level alignment is still not perfect. Thus, we consider further reducing class-level misalignment between the two domains. 

% \subsubsection{Entropy Minimization}
% {\color{green}
% CRA can also be interpreted from the perspective of entropy minimization \cite{}. Aligning the distributions of the untrusted (i,e., high-entropy) and trusted (i.e., low-entropy) pixels leads to reducing the overall entropy of the predicted classes at each pixel. As shown in several previous studies~\cite{advent,intrada,iast}, constraining the model output to low entropy is beneficial for the domain adaptation. We will show an experimental result verifying this interpretation in Table~\ref{tab:ab_sl_minent}.
% }
% Our experiments in Table~\ref{tab:ab_sl_minent} also verified this point, in which the unsupervised entropy minimization and hard domain label based CRA also help the final performance. 

{\color{black}

\subsection{Details of the Method} \label{ch:method}
%Alignment across Image Regions}
\begin{figure*}[ht]
\centering
\includegraphics[width=0.85\linewidth]{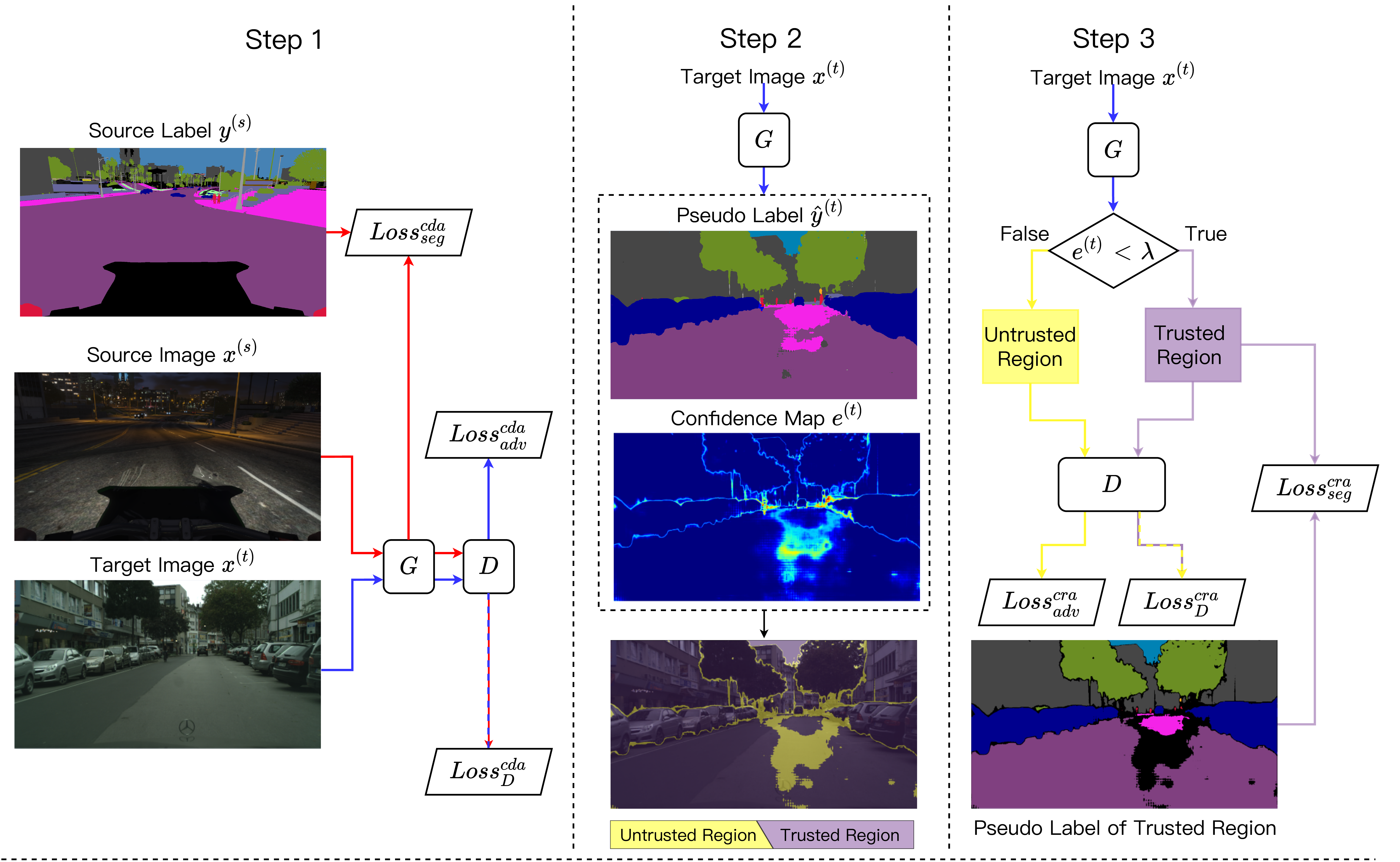}
\caption{Overall procedure of training a model ($G$) using labeled source domain data $\{(x^{(s)},y^{(s)})\}$ and unlabeled target domain data $\{x^{(t)}\}$. After training $G$ on the source domain data, we first apply a CDA method to align the feature distributions of the two domains, yielding updated $G$ and $D$ (Step 1). Next, for the target domain data, we generate pseudo labels $\hat{y}^{(t)}$ and split each image into trusted and untrusted regions based on a confidence map $e^{(t)}$  (Step 2). We finally apply the proposed CRA training to align the feature distributions of the two regions within the target domain data, resulting in updated $G$ (and $D$). 
}
%
% Overview of our proposed three-step approach. In the first step, we will use some existing methods to train the segmentation model $G$ with source domain data and reduce the gap between the source domain and the target domain via adversarial training. Then, we will generate pseudo labels $\hat{y}^{(t)}$ and entropy maps $e^{(t)}$ for target domain data to filter the trusted regions and the untrusted regions. We will conduct the region alignment in the final step, in which we treat the trusted regions as the source domain data and the untrusted regions as the target domain data and conduct adversarial training. This training scheme will boost the segmentation performance of the untrusted regions.} 
\label{pic:method}
\end{figure*}

% {\color{green} 
% Although the above method contributes to performance improvement, the class-level alignment is still not perfect. Thus, we consider further reducing class-level misalignment between the two domains. 
% }

\subsubsection{Choosing Trusted and Untrusted Image Regions.}
% Label Generation and Regions Split}

We first apply an existing CDA method and obtain a segmentation network $G$. We then classify each image pixel into two classes, {\em trusted} and {\em untrusted} using the entropy of the predicted class probability. Specifically we first apply $G$ to each image $x^{(t)}$ of the target domain, obtaining the softmax probability $p^{(t)}=[p_{i1}^{(t)},\ldots,p_{iK}^{(t)}]$ at  pixel $i$ of $x^{(t)}$. We calculate the entropy of the class probability as
\begin{equation}\label{eq:t_entropy_map}
e_{i} = -\frac{1}{K\log K}\sum_{k'=1}^{K}p_{ik'}^{(t)}\log(p_{ik'}^{(t)}).
\end{equation}
We then classify each pixel $(i)$ of the image by thresholding $e_i$ with a constant $\lambda$ into two classes, {\em trusted} and {\em untrusted}.
Let $m_{i}$ be a pixel-wise mask indicating the pixel being trusted, which is given by:
% $M_t =\{m_{jk}^{(tr)}, m_{jk}^{(utr)}\}_{j=1}^{n_t}$ 
\begin{equation}\label{eq:mask_trusted_region}
% m_{ji}^{(tr)} =
m_{i} =
\begin{cases}
    1 &\parbox{.14\textwidth}{if $ e_{i} < \lambda $} \\
    0 &\parbox{0.14\textwidth}{otherwise.}
\end{cases}
\end{equation}
The mask for untrusted pixels is obtained as $\overline{m}_i=1-m_i$. 
% \begin{equation}\label{eq:mask_untrusted_region}
% m_{ji}^{(utr)} =
% \begin{cases}
%     0, &\parbox{.14\textwidth}{if $ e_{i}^{(t)} \leq \lambda $} \\
%     1, &\parbox{0.14\textwidth}{otherwise}
% \end{cases}
% \end{equation}

% We apply adversarial training to align the feature distributions for the trusted and untrusted pixels. 
The choice of the hyperparameter $\lambda$ is important. If we have an access to validation data, as is assumed so in previous studies, we should choose it using them. Besides, there is a simple method to calculate a good value for $\lambda$. The minimum value for the entropy is given when  a single class has a probability $=1$, and the theoretical maximum is given when all the class probabilities are equal, i.e., $p_k=1/K$ $(k=1,\ldots,K)$, where $K$ is the number of classes. Now, suppose that the prediction is completely split among multiple classes and thus we cannot choose a particular class. The minimum entropy is given when two classes share the probability $=1/2$ and the others have zero, which is $\log 2/(K\log K)$. For the Cityscapes dataset, which has $K=19$ classes, this yields 0.012. The threshold $\lambda$ should be lower than this value; we choose $\lambda=0.01$ in our experiments. 

There is always class imbalance in segmentation data. Classes occupying less than a few percentage in the data tend to always yield high entropy and their pixels are mostly be treated as untrusted. To avoid this to occur, it is necessary to adaptively choose the threshold $\lambda$ for those classes. We found that a simple remedy works instead, which is to multiply the entropy for rare classes by $1/2$. The rare classes are identified by counting the number of pseudo labels $\hat{y}^{(t)}$ explained below and thresholding it with 1/100 of all the pixels. 

% {\color{blue} To avoid the situation where there are too few trusted regions to be selected, we will detect rare classes automatically and scale the entropy values of these classes during the CRA training. We will discuss more about this point in Sec.~\ref{ch:ab_ent_scaling}.}

% As mentioned before, the entropy of prediction probabilities can indicate whether the model is sure of its predictions. So, we can divide pixels into trusted and untrusted clusters based on the prediction entropy maps.

We will also use pseudo label for each pixel of a target domain image $x^{(t)}$. We define this as
% For a target domain sample $x^{(t)}$, whose pixel level labels are not available, we can generate the corresponding pseudo label $\hat{y}^{(t)}$ and entropy map $e^{(t)}$ using trained $G$ in step 1:
\begin{equation}\label{eq:t_pseudo_label}
\hat{y}_{ik}^{(t)} =
\begin{cases}
    1 &\parbox{0.2\textwidth}{if $k = \operatorname*{arg\,max}_{k'} p_{ik'}^{(t)}$} \\
    0 &\parbox{0.2\textwidth}{otherwise.}
\end{cases}
\end{equation}
}
% where $p^{(t)}$ is the softmax probability output of $G$, and $K$ represents the number of classes of the dataset. 
% In our generated $M_t$, for those pixels with prediction entropy below $\lambda$, we categorize them as trusted regions, and for pixels in the untrusted regions, their prediction entropy values are above the threshold $\lambda$.

\subsubsection{Training of the Network.}
% Cross Region Adaptation (CRA)} 
% \label{ch:cra}

% {\color{green} 
% We then align the feature distributions of the trusted and untrusted pixels within the target domain. We apply adversarial training to these distributions. It is similar to the standard adversarial training for UDA; here, the feature distribution for the trusted pixels is treated as the source domain distribution and that for the untrusted pixels is treated as the target domain distribution. 
% }

We train the segmentation network $G=C\circ F$ and the discriminator $D$ by fine-tuning the model trained in the previous step. 
We first train $G$ with the trusted pixels of the target domain images using the above pseudo label $\hat{y}_{ik}^{(t)}$. We ignore the untrusted pixels here. We use the standard cross-entropy loss:
\begin{equation}\label{eq:seg_loss_cra}
\mathcal{L}_{seg}^{cra} = -\sum\limits_{i=1}^{H \times W}\sum\limits_{k=1}^{K}m_i\hat{y}_{ik}^{(t)}\log p_{ik}^{(t)}.
\end{equation}
The insertion of the mask $m_i$ ensures the loss is computed over only the trusted pixels. 

% After this, we train $G$ and $D$ in an adversarial fashion, where $G$ and $D$ are updated alternately as follows. We first train $G$ so that $D$ will missclassify untrusted pixels as trusted pixels. We also employ the method of FADA, i.e., calculating the class information $a_{ik}^{(t)}$ and $a_{ik}^{(s)}$ according to (\ref{eq:def_soft_d_label}) and use them for the training of $G$ and $D$. Specifically, we minimize the following with respect to $G$ while freezing $D$.
After this, we train $G$ and $D$ in an adversarial fashion, where $G$ and $D$ are updated alternately as follows. We first train $G$ so that $D$ will missclassify untrusted pixels as trusted pixels. We employ the fine-grained adversarial approach~\cite{fada} for the training of $G$ and $D$, \textcolor{black}{in which we first generate a domain encoding label containing class probabilities, i.e., $[a;\mathbf{0}]$ for source domain and $[\mathbf{0};a]$ for target domain, where $a$ is a $K$-vector containing the class probabilities and $\mathbf{0}$ is a zero vector of size $K$. These are computed from the outputs of the segmentation network $G$. To be specific, $a$ is defined for pixel $i$ as follows:
\begin{equation}\label{eq:def_soft_d_label}
a_{ik} = \frac{\exp(\frac{z_{ik}}{T})}{\sum_{j=1}^{K}\exp(\frac{z_{ij}}{T})},
\end{equation}
where $z_{ik}$ is the logit for class $k$ from $G$ and $T$ is a hyperparameter (temperature). We denote the label by $[a^{(s)};a^{(t)}]$ below.} Then, we minimize the following with respect to $G$ while freezing $D$.
\begin{equation}\label{eq:adv_loss_cra}
\mathcal{L}_{adv}^{cra} =  
-\sum\limits_{i=1}^{H \times W}\sum\limits_{k=1}^{K}a_{ik}^{(t)}\overline{m}_i\log P(d=0,c=k \mid f_i). 
\end{equation}

The mask $\overline{m}_i$ ensures the sum is taken over the untrusted pixels. We then train $D$ to classify the input feature $f_i$ as trusted or untrusted regions as accurately as possible. This is done by minimizing
% Finally, we can optimize the segmentation model $G$ and the discriminator $D$ via adversarial training and domain soft labels for trusted regions and untrused regions, the training loss function can be written as:
\begin{multline}\label{eq:d_loss_cra}
\mathcal{L}_{D}^{cra} = 
-\sum\limits_{i=1}^{H \times W}\sum\limits_{k=1}^{K}a_{ik}^{(s)}m_i\log P(d=0,c=k \mid f_i) \\
-\sum\limits_{i=1}^{H \times W}\sum\limits_{k=1}^{K}a_{ik}^{(t)}\overline{m}_i\log P(d=1,c=k \mid f_i).
\end{multline}
As above, $m_i$ and $\overline{m}_i$ ensure the two sums taken over trusted and untrusted pixels, repectively. 

% In the entropy-based regions alignment stage, the trusted regions and the untrusted regions are from the same target domain samples. After training, the gap between different regions will be bridged.

In summary, after obtaining the trusted and untrusted regions we perform the following with a single minibatch and repeat it for a certain number of iterations: 
\begin{itemize}
    \item[1.] Update $G$ using $\mathcal{L}_{seg}^{cra}$ of (\ref{eq:seg_loss_cra}) with the pseudo labels (\ref{eq:t_pseudo_label}) on the trusted regions. 
    \item[2.] Compute the domain encoding label with class probabilities by (\ref{eq:def_soft_d_label}) and obtain $[0;a^{(t)}]$ for the untrusted region data and $[a^{(s)};0]$ for the trusted region data. 
    % This step is omitted when used with CRA methods other than FADA; hard domain labels are used instead below.
    \item[3.] Update $G$ using $\mathcal{L}^{cra}_{adv}$ of (\ref{eq:adv_loss_cra}). 
    \item[4.] Update $D$ using $\mathcal{L}^{cra}_D$ of (\ref{eq:d_loss_cra}).
\end{itemize}

\subsubsection{Overall Procedure.}

Given the labeled source domain data $X_s$ and unlabeled target domain data $X_t$, we perform the following steps to obtain a model ($G$) that performs well on the target domain. There are four steps in total. 
% The first is a warm-up step, where 
We start with training $G$ on $X_s$ in a standard supervised fashion. We then perform CDA 
% for aligning the feature distributions of the two domains
% , in which we use FADA~\cite{fada} (Sec.~\ref{ch:cda}), or
with any UDA method, to fine-tune $G$ and train $D$; see Step 1 of Fig.~\ref{pic:method}. Next, we generate pseudo labels and confidence maps for the images in $X_t$ and split each of them into trusted and untrusted regions according to the confidence maps, as in Sec.~\ref{ch:method}; see Step 2 of Fig.~\ref{pic:method}. Finally, we perform the CRA between the trusted and untrusted regions to fine-tune $G$ and $D$, as explained above;
% in Sec.~\ref{ch:cra}; 
see Step 3 of Fig.~\ref{pic:method}.

%-------------------------------------------------------------------------
\section{Experiments}

\begin{table*}[!htbp]
% \caption{Results for GTA5 $\rightarrow$ Cityscapes in terms of mIoU. The methods with the name `*+CRA' indicate the combination of a CDA method with the proposed CRA training. B is for backbones, V and R means VGG-16 and ResNet-101 backbones respectively. \textcolor{red}{$\Delta$ shows the improvement from the baseline CDA}. The best result of every class is highlighted. }
\caption{Results of UDA from (a) GTA5, (b) SYNTHIA, and (c) Synscapes to Cityscapes. `*+CRA' in the `method' column represents the combination of a CDA method and the proposed CRA. The column `B' indicates backbones; V and R means VGG-16 and ResNet-101, respectively. `($\Delta$)' in the last column indicates the improvement from the base CDA. The best result is highlighted for each class. The mIoU$_{16}$ and mIoU$_{13}$ in (b) denote the mIoU scores over 16 and 13 classes, respectively.}
\centering
\vspace*{-3mm}
{\scriptsize (a) GTA5 $\rightarrow$ Cityscapes}
\begin{adjustbox}{width=1\textwidth}
\begin{tabular}{l|c|ccccccccccccccccccc|c}
\hline
Method & B & \rotatebox{90}{road}& \rotatebox{90}{sidewalk}& \rotatebox{90}{building}& \rotatebox{90}{wall}& \rotatebox{90}{fence}& \rotatebox{90}{pole}& \rotatebox{90}{light}& \rotatebox{90}{sign}& \rotatebox{90}{veg}& \rotatebox{90}{terrain}& \rotatebox{90}{sky}& \rotatebox{90}{person}& \rotatebox{90}{rider}& \rotatebox{90}{car}& \rotatebox{90}{truck}& \rotatebox{90}{bus}& \rotatebox{90}{train}& \rotatebox{90}{mbike}& \rotatebox{90}{bike}& mIoU ($\Delta$)\\
\hline\hline
Source Only & V & 35.4 & 13.2 & 72.1 & 16.7 & 11.6 & 20.7 & 22.5 & 13.1 & 76.0 & 7.6 & 66.1 & 41.1 & 19.0 & 69.8 & 15.2 & 16.3 & 0.0 & 16.2 & 4.7 & 28.3 \\
AdaptSegNet~\cite{adaptseg} & V & 87.3 & 29.8 & 78.6 & 21.1 & 18.2 & 22.5 & 21.5 & 11.0 & 79.7 & 29.6 & 71.3 & 46.8 & 6.5 & 80.1 & 23.0 & 26.9 & 0.0 & 10.6 & 0.3 & 35.0 \\
ADVENT~\cite{advent} & V & 86.9 & 28.7 & 78.7 & 28.5 & 25.2 & 17.1 & 20.3 & 10.9 & 80.0 & 26.4 & 70.2 & 47.1 & 8.4 & 81.5 & 26.0 & 17.2 & 18.9 & 11.7 & 1.6 & 36.1 \\
FADA~\cite{fada} & V & 92.3 & 51.1 & 83.7 & 33.1 & 29.1 & 28.5 & 28.0 & 21.0 & 82.6 & 32.6 & 85.3 & 55.2 & 28.8 & 83.5 & 24.4 & 37.4 & 0.0 & 21.1 & 15.2 & 43.8\\
\rowcolor{highlight}
FADA+CRA & V & 93.3 & 59.3 & 84.6 & 24.8 & 26.1 & 36.2 & 33.5 & 30.2 & 84.2 & 38.4 & 85.6 & 60.4 & 29.6 & 85.0 & 24.7 & 39.0 & 20.4 & 24.3 & 17.5 & 47.2 (+3.4) \\
\hline
Source Only & R & 65.0 & 16.1 & 68.7 & 18.6 & 16.8 & 21.3 & 31.4 & 11.2 & 83.0 & 22.0 & 78.0 & 54.4 & 33.8 & 73.9 & 12.7 & 30.7 & 13.7 & 28.1 & 19.7 & 36.8\\
AdaptSegNet~\cite{adaptseg} & R & 86.5 & 36.0 & 79.9 & 23.4 & 23.3 & 23.9 & 35.2 & 14.8 & 83.4 & 33.3 & 75.6 & 58.5 & 27.6 & 73.7 & 32.5 & 35.4 & 3.9 & 30.1 & 28.1 & 42.4\\
\rowcolor{highlight}
AdaptSegNet+CRA & R & 87.7 & 33.5 & 81.6 & 33.3 & 23.7 & 22.4 & 26.1 & 16.0 & 83.6 & 38.5 & 72.7 & 52.7 & 21.2 & 84.2 & 44.9 & 43.0 & 0.0 & 26.2 & 33.0 & 43.4 (+1.0)\\
ADVENT~\cite{advent} & R & 89.9 & 36.5 & 81.6 & 29.2 & 25.2 & 28.5 & 32.3 & 22.4 & 83.9 & 34.0 & 77.1 & 57.4 & 27.9 & 83.7 & 29.4 & 39.1 & 1.5 & 28.4 & 23.3 & 43.8 \\ 
IntraDA~\cite{intrada} & R & 90.6 & 36.1 & 82.6 & 29.5 & 21.3 & 27.6 & 31.4 & 23.1 & 85.2 & 39.3 & 80.2 & 59.3 & 29.4 & 86.4 & 33.6 & 53.9 & 0.0 & 32.7 & 37.6 & 46.3 \\ 
\rowcolor{highlight}
ADVENT+CRA & R & 90.0 & 39.9 & 83.5 & 33.3 & 27.5 & 28.7 & 35.0 & 27.6 & 85.2 & 37.2 & 80.0 & 57.9 & 29.4 & 85.1 & 39.3 & 44.2 & 0.0 & 26.4 & 37.6 & 46.7 (+2.9)\\ 
FADA~\cite{fada} & R & 91.0 & 50.6 & 86.0 & 43.4 & 29.8 & 36.8 & 43.4 & 25.0 & 86.8 & 38.3 & 87.4 & 64.0 & 38.0 & 85.2 & 31.6 & 46.1 & 6.5 & 25.4 & 37.1 & 50.1\\
\rowcolor{highlight}
FADA+CRA & R & 91.6 & 53.6 & 85.5 & 42.6 & 18.7 & 34.8 & 36.3 & 18.9 & 87.8 & 45.0 & \textbf{89.0} & 66.2 & 39.0 & 87.6 & 42.3 & 51.3 & 14.9 & 42.0 & 44.6 & 52.2 (+2.1) \\
IAST~\cite{iast} & R & 94.1 & 58.8 & 85.4 & 39.7 & 29.2 & 25.1 & 43.1 & 34.2 & 84.8 & 34.6 & 88.7 & 62.7 & 30.3 & 87.6 & 42.3 & 50.3 & \textbf{24.7} & 35.2 & 40.2 & 52.2 \\
\rowcolor{highlight}
IAST+CRA & R & \textbf{93.3} & 57.8 & \textbf{87.2} & 42.1 & 29.6 & 40.9 & 50.8 & 51.0 & 86.0 & 28.2 & 87.7 & 67.8 & 34.4 & 87.5 & 29.8 & 46.3 & 14.2 & 40.8 & 52.7 & 54.1 (+1.9)\\
ProDA~\cite{proda} & R & 87.8 & 56.0 & 79.7 & 46.3 & 44.8 & \textbf{45.6} & 53.5 & 53.5 & 88.6 & 45.2 & 82.1 & 70.7 & 39.2 & \textbf{88.8} & \textbf{45.5} & \textbf{59.4} & 1.0 & 48.9 & 56.4 & 57.5 \\
\rowcolor{highlight}
ProDA+CRA & R & 89.4 & \textbf{60.0} & 81.0 & \textbf{49.2} & \textbf{44.8} & 45.5 & \textbf{53.6} & \textbf{55.0} & \textbf{89.4} & \textbf{51.9} & 85.6 & \textbf{72.3} & \textbf{40.8} & 88.5 & 44.3 & 53.4 & 0.0 & \textbf{51.7} & \textbf{57.9} & \textbf{58.6} (+1.1)\\
\hline 
\end{tabular}
\end{adjustbox}
{\scriptsize(b) SYNTHIA $\rightarrow$ Cityscapes}
\begin{adjustbox}{width=1\textwidth}
\begin{tabular}{l|c|cccccccccccccccc|c|c}
\hline
Method & B & \rotatebox{90}{road}& \rotatebox{90}{sidewalk}& \rotatebox{90}{building}& \rotatebox{90}{wall}& \rotatebox{90}{fence}& \rotatebox{90}{pole}& \rotatebox{90}{light}& \rotatebox{90}{sign}& \rotatebox{90}{veg}& \rotatebox{90}{sky}& \rotatebox{90}{person}& \rotatebox{90}{rider}& \rotatebox{90}{car}& \rotatebox{90}{bus}& \rotatebox{90}{mbike}& \rotatebox{90}{bike}& mIoU$_{16}$ ($\Delta$)& mIoU$_{13}$ ($\Delta$)\\
\hline\hline
Source Only & V & 10.0 & 14.7 & 52.4 & 4.2 & 0.1 & 20.9 & 3.5 & 6.5 & 74.3 & 77.5 & 44.9 & 4.9 & 64.0 & 21.6 & 4.2 & 6.4 & 25.6 & 29.6\\ 
AdaptSegNet~\cite{adaptseg} & V & 78.9 & 29.2 & 75.5 & - & - & - & 0.1 & 4.8 & 72.6 & 76.7 & 43.4 & 8.8 & 71.1 & 16.0 & 3.6 & 8.4 & - & 37.6 \\
ADVENT~\cite{advent} & V & 67.9 & 29.4 & 71.9 & 6.3 & 0.3 & 19.9 & 0.6 & 2.6 & 74.9 & 74.9 & 35.4 & 9.6 & 67.8 & 21.4 & 4.1 & 15.5 & 31.4 & 36.6 \\
FADA~\cite{fada} & V & 80.4 & 35.9 & 80.9 & 2.5 & 0.3 & 30.4 & 7.9 & 22.3 & 81.8 & 83.6 & 48.9 & 16.8 & 77.7 & 31.1 & 13.5 & 17.9 & 39.5 & 46.0\\
\rowcolor{highlight}
FADA+CRA & V & 83.7 & 38.7 & 79.4 & 0.3 & 0.6 & 30.6 & 0.0 & 8.5 & 81.8 & 78.1 & 58.7 & 17.4 & 80.7 & 32.8 & 14.6 & 45.3 & 40.7 (+1.2) & 47.7 (+1.7)\\
\hline
Source Only & R & 55.6 & 23.8 & 74.6 & 9.2 & 0.2 & 24.4 & 6.1 & 12.1 & 74.8 & 79.0 & 55.3 & 19.1 & 39.6 & 23.3 & 13.7 & 25.0 & 33.5 & 38.6 \\
AdaptSegNet~\cite{adaptseg} & R & 84.3 & 42.7 & 77.5 & - & - & - & 4.7 & 7.0 & 77.9 & 82.5 & 54.3 & 21.0 & 72.3 & 32.2 & 18.9 & 32.3 & - & 46.7 \\ 
ADVENT~\cite{advent} & R & 85.6 & 42.2 & 79.7 & 8.7 & 0.4 & 25.9 & 5.4 & 8.1 & 80.4 & 84.1 & 57.9 & 23.8 & 73.3 & 36.4 & 14.2 & 33.0 & 41.2 & 48.0 \\
IntraDA~\cite{intrada} & R & 84.3 & 37.7 & 79.5 & 5.3 & 0.4 & 24.9 & 9.2 & 8.4 & 80.0 & 84.1 & 57.2 & 23.0 & 78.0 & 38.1 & 20.3 & 36.5 & 41.7 & 48.9\\
FADA~\cite{fada} & R & 84.5 & 40.1 & 83.1 & 4.8 & 0.0 & 34.3 & 20.1 & 27.2 & 84.8 & 84.0 & 53.5 & 22.6 & 85.4 & 43.7 & 26.8 & 27.8 & 45.2 & 52.5 \\
\rowcolor{highlight}
FADA+CRA & R & \textbf{88.8} & \textbf{48.3} & 83.4 & 12.6 & 0.4 & 36.4 & 8.1 & 22.2 & 86.6 & 80.9 & 64.8 & 22.4 & 87.7 & 45.5 & 29.9 & 46.4 & 47.8 (+2.6) & 55.0 (+2.5) \\
IAST~\cite{iast} & R & 81.9 & 41.5 & 83.3 & 17.7 & \textbf{4.6} & 32.3 & 30.9 & 28.8 & 83.4 & 85.0 & 65.5 & 30.8 & 86.5 & 38.2 & 33.1 & 52.7 & 49.8 & 57.0 \\
\rowcolor{highlight}
IAST+CRA & R & 75.8 & 33.3 & \textbf{85.1} & 36.7 & 0.0 & 42.5 & 46.7 & 37.2 & 85.4 & 83.7 & 68.5 & \textbf{32.0} & 87.8 & 44.9 & 40.7 & 53.8 & 53.4 (+3.6) & 59.6 (+2.6)\\
ProDA~\cite{proda} & R & 87.8 & 45.7 & 84.6 & 37.1 & 0.6 & \textbf{44.0} & 54.6 & 37.0 & \textbf{88.1} & 84.4 & 74.2 & 24.3 & 88.2 & 51.1 & 40.5 & 45.6 & 55.5 & 62.0 \\
\rowcolor{highlight}
ProDA+CRA & R & 85.6 & 44.2 & 82.7 & \textbf{38.6} & 0.4 & 43.5 & \textbf{55.9} & \textbf{42.8} & 87.4 & \textbf{85.8} & \textbf{75.8} & 27.4 & \textbf{89.1} & \textbf{54.8} & \textbf{46.6} & \textbf{49.8} & \textbf{56.9} (+1.4) & \textbf{63.7} (+1.7)\\
\hline
\end{tabular}
\end{adjustbox}
{\scriptsize(c) Synscapes $\rightarrow$ Cityscapes}
\begin{adjustbox}{width=1\textwidth}
\begin{tabular}{l|c|ccccccccccccccccccc|c}
\hline
Method & B & \rotatebox{90}{road}& \rotatebox{90}{sidewalk}& \rotatebox{90}{building}& \rotatebox{90}{wall}& \rotatebox{90}{fence}& \rotatebox{90}{pole}& \rotatebox{90}{light}& \rotatebox{90}{sign}& \rotatebox{90}{veg}& \rotatebox{90}{terrain}& \rotatebox{90}{sky}& \rotatebox{90}{person}& \rotatebox{90}{rider}& \rotatebox{90}{car}& \rotatebox{90}{truck}& \rotatebox{90}{bus}& \rotatebox{90}{train}& \rotatebox{90}{mbike}& \rotatebox{90}{bike}& mIoU ($\Delta$)\\
\hline\hline
Source Only & R & 81.8 & 40.6 & 76.1 & 23.3 & 16.8 & 36.9 & 36.8 & 40.1 & 83.0 & 34.8 & 84.9 & 59.9 & 37.7 & 78.4 & 20.4 & 20.5 & 7.8 & 27.3 & 52.5 & 45.3 \\
AdaptSegNet~\cite{adaptseg} & R & 94.2 & 60.9 & 85.1 & 29.1 & 25.2 & 38.6 & 43.9 & 40.8 & 85.2 & 29.7 & 88.2 & 64.4 & 40.6 & 85.8 & 31.5 & 43.0 & 28.3 & 30.5 & 56.7 & 52.7 \\
FADA & R & 94.1 & 58.4 & 84.4 & 33.7 & 34.7 & 38.4 & 44.0 & 43.2 & 84.7 & 40.6 & 89.2 & 62.9 & 38.8 & 87.0 & 29.3 & 40.2 & 18.2 & 33.1 & 55.8 & 53.2 \\
IntraDA~\cite{intrada} & R & 94.0 & 60.0 & 84.9 & 29.5 & 26.2 & 38.5 & 41.6 & 43.7 & 85.3 & 31.7 & 88.2 & 66.3 & 44.7 & 85.7 & 30.7 & 53.0 & 29.5 & 36.5 & 60.2 & 54.2 \\
\rowcolor{highlight}
FADA+CRA & R & 94.2 & 58.9 & 85.8 & 33.9 & 37.8 & 41.7 & 47.6 & 46.3 & 85.6 & 45.4 & 90.3 & 66.8 & 40.0 & 87.1 & 31.5 & 43.5 & 18.9 & 37.6 & 59.5 & 55.3 (+2.1) \\
IAST & R & 94.0 & 64.4 & 85.7 & 34.2 & 35.2 & 47.4 & 51.0 & 61.1 & 88.1 & 42.0 & \textbf{90.5} & 70.4 & 35.3 & 86.7 & 28.3 & 37.0 & 23.1 & 32.8 & 52.5 & 55.8 \\
\rowcolor{highlight}
IAST+CRA & R & \textbf{94.9} & \textbf{68.0} & \textbf{87.2} & \textbf{37.4} & \textbf{39.5} & 48.5 & 50.6 & 64.3 & 88.2 & 39.9 & 88.9 & 72.7 & 36.4 & 86.2 & 29.8 & 37.7 & 18.2 & 31.6 & 54.4 & 56.5 (+0.7) \\
ProDA & R & 90.7 & 43.1 & 85.3 & 29.5 & 32.8 & 50.1 & 61.8 & \textbf{64.5} & 89.4 & \textbf{48.6} & 89.1 & 76.3 & 47.0 & 88.9 & \textbf{36.5} & 54.4 & \textbf{34.8} & \textbf{44.4} & 61.0 & 59.4 \\
\rowcolor{highlight}
ProDA+CRA & R & 90.1 & 42.9 & 85.0 & 33.4 & 36.7 & \textbf{52.2} & \textbf{63.7} & 64.0 & \textbf{90.1} & 48.5 & 89.6 & \textbf{77.6} & \textbf{49.4} & \textbf{89.3} & 36.1 & \textbf{56.6} & 32.9 & 44.3 & \textbf{61.3} & \textbf{60.2} (+0.8)\\
\hline
\end{tabular}
\end{adjustbox}
\label{tab:results}
\end{table*}

\begin{figure*}[!htbp]
\centering
\includegraphics[width=0.98\linewidth]{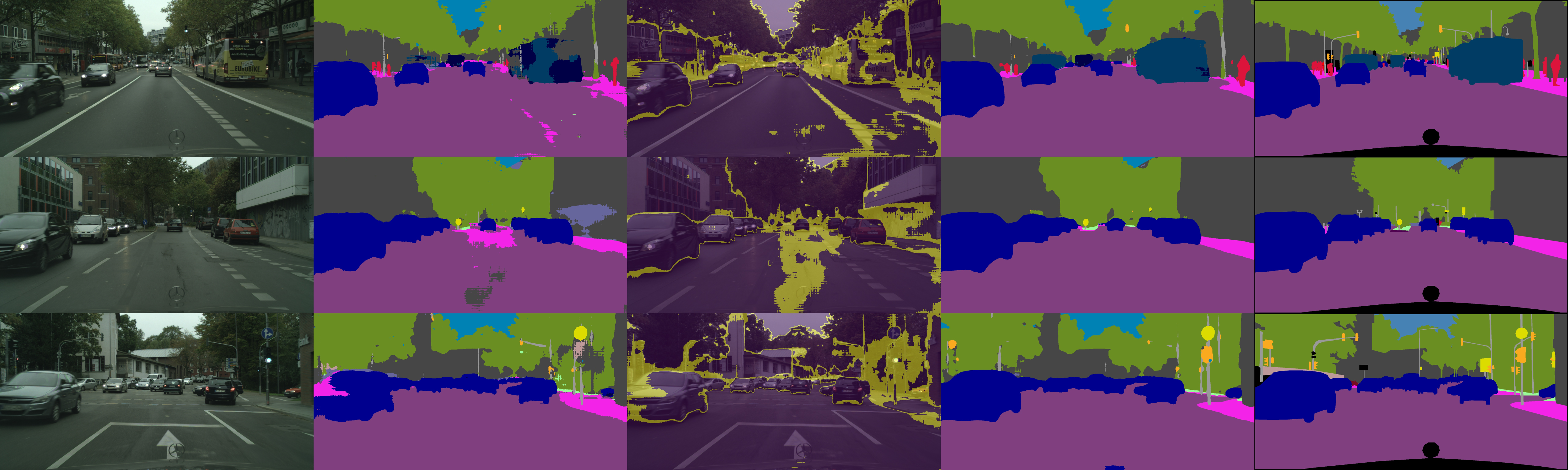}
\caption{Examples of the results on GTA5 $\rightarrow$ Cityscapes. From left to right, target images, results of CDA (FADA \cite{fada}), trusted and untrusted regions (in purple and yellow, respectively), results of CRA, and ground truths. }
% In region splitting part, we use light yellow to indicate the untrusted regions and light purple to indicate the trusted regions.} 
\label{pic:samples}
\end{figure*}

%-------------------------------------------------------------------------
\subsection{Datasets}

We evaluate our method on three scenarios of domain adaptation from a synthetic to a real dataset. For the source domain dataset, we consider either of GTA5 \cite{gta5}, SYNTHIA \cite{synthia}, and Synscapes \cite{synscapes}. For the target domain dataset, we consider  Cityscapes~\cite{cityscapes}.

\medskip
\noindent \textbf{Cityscapes} \cite{cityscapes}~~ This is an urban scene dataset consisting of data collected from the real world, which contains 19 categories. It provides 2,975 images for training, 500 images for validation, and 1,525 images for testing. For the use of these data, we follow previous studies \cite{advent,adaptseg,fada}. We report the performance on the validation set below unless otherwise noted. We also report the results on the test set in Sec.~\ref{ch:ab_online_test}.

\medskip
\noindent \textbf{GTA5} \cite{gta5}~~ This is a synthetic dataset generated from a video game. It contains 24,966 images with segmentation labels and shares the same 19 classes as Cityscapes. We use all the images as the source training data.

\medskip
\noindent \textbf{SYNTHIA} \cite{synthia}~~ This is a synthetic dataset which mainly contains urban scene samples. In our experiments, we use the SYNTHIA-RAND-CITYSCAPES subset as our source domain training data, which shares 16 classes with Cityscapes and has 9,400 images with segmentation labels.

\medskip
\noindent \textbf{Synscapes} \cite{synscapes}~~ This  is a photorealistic synthetic dataset for street scene parsing. It shares the same 19 classes as Cityscapes and contains 25,000 images with segmentation labels.

%-------------------------------------------------------------------------
\subsection{Experimental Configuration}
%-------------------------------------------------------------------------
% \subsection{Implementation Details}

For the design of the segmentation network $G$ and the discriminator $D$, we 
follow previous studies.
% of semantic segmentation UDA. 
To be specific, we use 
DeepLabv2~\cite{deeplab} with two different backbone (i.e., feature extraction networks),  VGG-16~\cite{vgg} and ResNet101~\cite{resnet}, for $G$. Each backbone is pretrained on ImageNet~\cite{imagenet}. We use a network consisting of three convolution layers for $D$. 
% For training procedure, 
% we follow the previous work \cite{fada}. Specifically, we first train $G$ on the source data for 20k iterations. We then perform CDA on $G$ and $D$ using an existing method for 40k iterations.
\textcolor{black}{
We train $G$ and $D$ as explained in Sec.~\ref{ch:method}. We first train $G$ on the source domain and then apply an existing CDA method for the alignment of the feature distributions across the domains, where we follow the experimental setting of the CDA method. We then apply the proposed CRA method for 40k iterations.}
% we first conduct the warm-up training and the CDA training according to the settings of every baseline, and then fine-tune $G$ and $D$ with the proposed CRA for 40k iterations.}

% To examine the effectiveness of CRA, 
For the CDA method, we select the existing state-of-the-art CDA methods, i.e., AdaptSegNet~\cite{adaptseg}, ADVENT~\cite{advent}, FADA~\cite{fada}, IAST~\cite{iast} and ProDA~\cite{proda}. The results obtained by the combination of a CDA and the proposed CRA will be referred to as `(the CDA)+CRA'; for example, ProDA+CRA indicates the combination of ProDA and the proposed CRA.

We use the SGD optimizer for the training of $G$ with momentum $=0.9$ and weight decay $=10^{-4}$. The learning rate follows polynomial decay from $2.5 \times 10^{-4}$ with power of 0.9. We use the Adam optimizer for the training of $D$; the initial learning rate is set to $10^{-4}$, and $\beta_1 = 0.9$ and $\beta_2 = 0.99$. We use the polynomial decay of learning rate. 

For evaluation, we follow the standard method for semantic segmentation~\cite{deeplab}; we use the intersection-over-union (IoU)~\cite{pascalvoc} as our metric. We report per-class IoU and mean IoU over all classes.
%-------------------------------------------------------------------------
\subsection{Main Results on Different UDA Scenarios}

\subsubsection{GTA5 $\rightarrow$ Cityscapes.}
Table~\ref{tab:results}(a) shows the results. ProDA+CRA, the combination of ProDA~\cite{proda} with CRA, achieves 58.6\% mean IoU, which outperforms all the previous methods. As compared with the original ProDA, it brings about 1.1pp for the ResNet-101 backbone. The combinations of CRA with other CDA methods, AdaptSegNet, ADVENT, FADA and IAST, also improve the baselines.
% For the results of IAST~\cite{iast}, which performed three rounds of training, we report the result obtained by one round of training for a fair comparison.
% with other methods on the task GTA5 $\rightarrow$ Cityscapes, FADA-CRA means the combination of FADA~\cite{fada} and CRA training. Overall, our proposed FADA-CRA can achieve 52.2\% mean IoU score, which outperforms previous methods. And compared to the vanilla FADA baseline, our CRA training can bring mean IoU improvement of 3.4pp and 2.1pp for the VGG-16-based FADA and ResNet-101-based FADA, respectively. These results demonstrate the effectiveness of proposed CRA training.
Fig.~\ref{pic:samples} shows a few examples of the results of FADA+CRA.
% These results visually illustrate the effectiveness of our methods. 
It is observed that the CDA (i.e., FADA) fails in most of the image regions classified as untrusted based on the entropy. It is also seen that the application of CRA consistently leads to the improved segmentation results. 

% First, for the region splitting, our entropy-based method is precise, most of the untrusted regions we find are the regions with wrong classifications. The second point is our proposed CRA training can consistently lead to more accurate segmentation results and yield finer details on the untrusted regions. For the ``bus" in the first row, the CRA training can classify pixels more accurate. In the second row, for the large scale stuff category ``road", our method can correct the misclassifications in the center of the image. And in the last row, our proposed CRA training can also deal with areas with many classes like the right side of the target image.

\subsubsection{SYNTHIA $\rightarrow$ Cityscapes.} Table \ref{tab:results}(b) shows the results. ProDA+CRA outperforms the original ProDA method and achieves the best performance. 

\subsubsection{Synscapes $\rightarrow$ Cityscapes.} Table \ref{tab:results}(c) shows the results. ProDA+CRA achieves the best mIoU score $=60.2$, which is 0.8pp higher than the former state-of-the-art \cite{proda}.

\subsection{Comparison with Entropy Minimization}
% Other Optimization Methods for Untrusted Regions}
% Different Optimization of Untrusted Regions} \label{ch:ab_sl_minent}

{\color{black} CRA aligns the feature distributions of the untrusted (i.e., high-entropy) regions and the trusted (low-entropy) regions. As it will effectively reduce the areas of the untrusted regions, the closest is the traditional entropy minimization. We conducted an experiment to compare CRA with the entroy minimization and a few other related methods. For the base CDA, we chose FADA without its two optional steps (i.e., self-distillation and multi-scale testing) \cite{fada}. 
% The proposed method optimize the untrusted regions via CRA for better performance. To examine the effectiveness of CRA, we experimentally compare it against other optimization methods in an equal condition. In the experiment, consider FADA without the above two optional steps as a base model. 
We first apply it to the target domain images,
% getting class probabilities at each image pixel and 
splitting the image into the two regions. We then apply two methods instead of CRA. The first is to retrain the base model using the pseudo labels obtained only for the trusted regions
% only on the image pixels with low entropy, which we call trusted pixels, using their pseudo labels 
\cite{zou2018unsupervised,zou2019confidence}. The second is the entropy minimization,
% to sharpen the prediction results of untrusted regions~
specifically {\color{black} adding a loss minimizing the entropy of untrusted regions to the loss of training on the pseudo labels.
% \cite{iast}}. 
% The proposed CRA employs a soft label for representing the domain label. We can use the hard label instead and tested its performance for comparison. 
% The third method performs adversarial training between the trusted and untrusted image regions with hard domain labels like standard adversarial training~\cite{adaptseg,advent}. 
Table \ref{tab:ab_sl_minent} shows the results. It is seen that CRA
% our method, whether using soft or hard labels, 
works better than the others, validating our approach.}

\begin{table}[H]
\caption{Comparison with entropy minimization etc.}
\centering
% \begin{adjustbox}{width=0.6\columnwidth}
\small
\begin{tabular}{l|c}
\hline
Method & mIoU \\
\hline
CDA (FADA w/o options) & 46.9 \\
Only training on pseudo labels & 47.5 \\
Entropy minimization & 48.4 \\
% CRA w/ Hard Domain Labels & 49.6 \\
% CRA w/ soft domain labels & \textbf{50.4} \\
CRA & \textbf{50.4} \\
% mIoU & 50.1 & 50.7 & 52.2 \\
\hline
\end{tabular}
% \end{adjustbox}
\label{tab:ab_sl_minent}
\end{table}

\subsection{Evaluation with the Cityscapes Test Set} \label{ch:ab_online_test}

The Cityscapes dataset is primarily used as the target domain dataset in existing studies. The dataset consists of training, validation, and test subsets. The ground truth labels for the test subset are unavailable, and the evaluation of results on it needs to post them to the official server \cite{deeplab,danet,encnet}. Thus, a common practice of the existing studies is to evaluate methods on the validation subset. However, this may make the fair comparison very hard, considering the necessity of choosing hyperparemters also on the validation subset. To cope with this, we evaluate the results for the test subset by the proposed method and the compared methods on the official server. Table \ref{tab:ab_online_test} shows the results, which validates the effectiveness of the proposed approach.

% 既存研究ではcityscapesがtarget domainとしてprimaryに使われている．cityscapesはtrain, val, testの3スプリットがあるが，testのground truthは非公開であり，その上での評価はofficial serverへのポストを要する．そのため，既存研究の大部分でval splitでの評価が行われている．しかしhyperparemterの選択を考えると，これは正しい評価とは言えない．そこでわれわれは，提案手法とその他の手法をofficial server上で評価してもらった．結果を表に示す．

% In the current evaluation procedure, especially when Cityscapes~\cite{cityscapes} is used as the target domain, existing methods often use the validation set of Cityscapes to make performance comparisons, and we think it is not a rigorous method. So we also evaluate the performance of some previous methods and our method with GTA5 $\rightarrow$ Cityscapes task and the test set of Cityscapes in Table~\ref{tab:ab_online_test}. The baseline FADA~\cite{fada} method can get 52.4 and our FADA-CRA is 53.8 in terms of mIoU score. If we further improve our performance by iterative training, our method can achieve 54.6 mIoU on the test set of Cityscapes. These results illustrate the effectiveness of our method.

\begin{table}[H]
\caption{Results on the test set of Cityscapes.}
\centering
% \begin{adjustbox}{width=0.65\columnwidth}
\small
\begin{tabular}{l|c}
\hline
% Method & IoU cla. & iIoU cla. & IoU cat. & iIoU cat. \\
Method & mIoU \\
\hline
% AdaptSegNet~\cite{adaptseg} & 43.6 & 16.6 & 73.4 & 42.4 \\
% ADVENT~\cite{advent} & 45.7 & 18.4 & 75.4 & 46.8 \\
% IntraDA~\cite{intrada} & 47.2 & 19.4 & 75.5 & 45.7 \\
% FADA~\cite{fada} & 52.4 & 25.6 & 79.3 & 58.3 \\
% IAST~\cite{iast} & 52.7 & 26.1 & 79.2 & 58.0 \\
% FADA-CRA & \textbf{53.8} & \textbf{27.5} & \textbf{79.4} & \textbf{58.3} \\
AdaptSegNet & 43.6 \\
ADVENT & 45.7 \\
IntraDA & 47.2 \\
FADA & 52.4 \\
% IAST & 52.7 \\
FADA+CRA & 53.8 \\
% IAST w/ Iterative Training & 54.4 \\
IAST & 54.4 \\
IAST+CRA & 55.6 \\
% FADA-CRA w/ Iterative Training & \textbf{54.6} \\
% FADA-CRA (IT) & \textbf{54.6} \\
ProDA & 56.9 \\
ProDA+CRA & \textbf{58.9} \\
\hline
\end{tabular}
% \end{adjustbox}
\label{tab:ab_online_test}
\end{table}

%-------------------------------------------------------------------------
\section{Summary and Conclusion}

We have presented a method for unsupervised domain adaptation for semantic segmentation, named cross-region adaptation (CRA). Its basic idea is to perform adversarial training inside the target domain, aiming to reduce class-level misalignment of feature distributions between the source and the target domains. The proposed method specifies the two feature distributions in the target domain in a self-training framework. After applying an existing UDA method, which we refer to as cross-domain adaptation (CDA), to best align two feature distributions from different domains, the proposed method splits each target domain image into trusted and untrusted regions based on a confidence map. It then aligns the feature distributions from the two image regions by adversarial training. 

% We have shown experimental results showing that the proposed CRA method combined with FADA achieves new state-of-the-art in the standard tests of adaptation from three different CG datasets to Cityscapes. When combined with other CDA methods, the proposed approach yields better performance than using the combined method alone. We can regard the proposed method as integration of adversarial training and self-training. Our experiments show that the integration is effective since it outperforms pure self-training methods. Finally, we compared the major UDA methods using the Cityscapes test dataset to aim at the fairest comparison without a leakage from training to test data. The results show that the proposed approach yields the best performance. All these results support the effectiveness of the proposed approach.
We have shown experimental results showing that the proposed CRA method combined with ProDA achieves new state-of-the-art in the standard tests of adaptation from three different CG datasets to Cityscapes. When combined with other CDA methods, the proposed approach yields better performance than using the combined method alone. 
% \textcolor{blue}{We can regard the proposed method as optimization for untrusted pixels. Our experiments show that our method is effective since it outperforms other optimization methods for untrusted pixels.} 
Finally, we compared the major UDA methods using the Cityscapes test dataset to aim at the fairest comparison without a leakage from training to test data. The results show that the proposed approach yields the best performance. All these results support the effectiveness of the proposed approach.

\bibliography{ref}

\begin{thebibliography}{38}
\providecommand{\natexlab}[1]{#1}

\bibitem[{Badrinarayanan, Kendall, and Cipolla(2017)}]{segnet}
Badrinarayanan, V.; Kendall, A.; and Cipolla, R. 2017.
\newblock SegNet: A Deep Convolutional Encoder-Decoder Architecture for Image
  Segmentation.
\newblock \emph{IEEE Transactions on Pattern Analysis and Machine
  Intelligence}, 39(12): 2481--2495.

\bibitem[{Baktashmotlagh et~al.(2013)Baktashmotlagh, Harandi, Lovell, and
  Salzmann}]{Baktashmotlagh_2013_ICCV}
Baktashmotlagh, M.; Harandi, M.~T.; Lovell, B.~C.; and Salzmann, M. 2013.
\newblock Unsupervised Domain Adaptation by Domain Invariant Projection.
\newblock In \emph{Proceedings of the IEEE/CVF International Conference on
  Computer Vision (ICCV)}, 769--776.

\bibitem[{Chen et~al.(2019{\natexlab{a}})Chen, Xie, Huang, Rong, Ding, Huang,
  Xu, and Huang}]{PFAN_2019_CVPR}
Chen, C.; Xie, W.; Huang, W.; Rong, Y.; Ding, X.; Huang, Y.; Xu, T.; and Huang,
  J. 2019{\natexlab{a}}.
\newblock Progressive Feature Alignment for Unsupervised Domain Adaptation.
\newblock In \emph{Proceedings of the IEEE/CVF Conference on Computer Vision
  and Pattern Recognition (CVPR)}, 627--636.

\bibitem[{Chen et~al.(2017)Chen, Papandreou, Kokkinos, Murphy, and
  Yuille}]{deeplab}
Chen, L.-C.; Papandreou, G.; Kokkinos, I.; Murphy, K.; and Yuille, A.~L. 2017.
\newblock DeepLab: Semantic Image Segmentation with Deep Convolutional Nets,
  Atrous Convolution, and Fully Connected CRFs.
\newblock \emph{IEEE Transactions on Pattern Analysis and Machine
  Intelligence}, 40(4): 834--848.

\bibitem[{Chen et~al.(2019{\natexlab{b}})Chen, Li, Chen, and
  Gool}]{chen2019learning}
Chen, Y.; Li, W.; Chen, X.; and Gool, L.~V. 2019{\natexlab{b}}.
\newblock Learning Semantic Segmentation from Synthetic Data: A Geometrically
  Guided Input-Output Adaptation Approach.
\newblock In \emph{Proceedings of the IEEE/CVF Conference on Computer Vision
  and Pattern Recognition (CVPR)}, 1841--1850.

\bibitem[{Cordts et~al.(2016)Cordts, Omran, Ramos, Rehfeld, Enzweiler,
  Benenson, Franke, Roth, and Schiele}]{cityscapes}
Cordts, M.; Omran, M.; Ramos, S.; Rehfeld, T.; Enzweiler, M.; Benenson, R.;
  Franke, U.; Roth, S.; and Schiele, B. 2016.
\newblock The Cityscapes Dataset for Semantic Urban Scene Understanding.
\newblock In \emph{Proceedings of the IEEE/CVF Conference on Computer Vision
  and Pattern Recognition (CVPR)}, 3213--3223.

\bibitem[{Deng et~al.(2009)Deng, Dong, Socher, Li, Li, and Fei-Fei}]{imagenet}
Deng, J.; Dong, W.; Socher, R.; Li, L.-J.; Li, K.; and Fei-Fei, L. 2009.
\newblock ImageNet: a Large-Scale Hierarchical Image Database.
\newblock In \emph{Proceedings of the IEEE/CVF Conference on Computer Vision
  and Pattern Recognition (CVPR)}, 248--255.

\bibitem[{Everingham et~al.(2015)Everingham, Eslami, Van~Gool, Williams, Winn,
  and Zisserman}]{pascalvoc}
Everingham, M.; Eslami, S.~A.; Van~Gool, L.; Williams, C.~K.; Winn, J.; and
  Zisserman, A. 2015.
\newblock The PASCAL Visual Object Classes Challenge: A Retrospective.
\newblock \emph{International Journal of Computer Vision (IJCV)}, 111(1):
  98--136.

\bibitem[{Fu et~al.(2019)Fu, Liu, Tian, Li, Bao, Fang, and Lu}]{danet}
Fu, J.; Liu, J.; Tian, H.; Li, Y.; Bao, Y.; Fang, Z.; and Lu, H. 2019.
\newblock Dual Attention Network for Scene Segmentation.
\newblock In \emph{Proceedings of the IEEE/CVF Conference on Computer Vision
  and Pattern Recognition (CVPR)}, 3146--3154.

\bibitem[{Gong et~al.(2019)Gong, Li, Chen, and Gool}]{gong2019dlow}
Gong, R.; Li, W.; Chen, Y.; and Gool, L.~V. 2019.
\newblock DLOW: Domain Flow for Adaptation and Generalization.
\newblock In \emph{Proceedings of the IEEE/CVF Conference on Computer Vision
  and Pattern Recognition (CVPR)}, 2477--2486.

\bibitem[{He et~al.(2016)He, Zhang, Ren, and Sun}]{resnet}
He, K.; Zhang, X.; Ren, S.; and Sun, J. 2016.
\newblock Deep Residual Learning for Image Recognition.
\newblock In \emph{Proceedings of the IEEE/CVF Conference on Computer Vision
  and Pattern Recognition (CVPR)}, 770--778.

\bibitem[{Lee et~al.(2018)Lee, Tseng, Huang, Singh, and Yang}]{lee2018diverse}
Lee, H.-Y.; Tseng, H.-Y.; Huang, J.-B.; Singh, M.; and Yang, M.-H. 2018.
\newblock Diverse Image-to-Image Translation via Disentangled Representations.
\newblock In \emph{Proceedings of the European Conference on Computer Vision
  (ECCV)}, 35--51.

\bibitem[{Li, Yuan, and Vasconcelos(2019)}]{li2019bidirectional}
Li, Y.; Yuan, L.; and Vasconcelos, N. 2019.
\newblock Bidirectional Learning for Domain Adaptation of Semantic
  Segmentation.
\newblock In \emph{Proceedings of the IEEE/CVF Conference on Computer Vision
  and Pattern Recognition (CVPR)}, 6936--6945.

\bibitem[{Long, Shelhamer, and Darrell(2015)}]{fcn}
Long, J.; Shelhamer, E.; and Darrell, T. 2015.
\newblock Fully Convolutional Networks for Semantic Segmentation.
\newblock In \emph{Proceedings of the IEEE/CVF Conference on Computer Vision
  and Pattern Recognition (CVPR)}, 3431--3440.

\bibitem[{Long et~al.(2015)Long, Cao, Wang, and Jordan}]{long2015learning}
Long, M.; Cao, Y.; Wang, J.; and Jordan, M. 2015.
\newblock Learning Transferable Features with Deep Adaptation Networks.
\newblock In \emph{Proceedings of the International Conference on Machine
  Learning}, 97--105.

\bibitem[{Mei et~al.(2020)Mei, Zhu, Zou, and Zhang}]{iast}
Mei, K.; Zhu, C.; Zou, J.; and Zhang, S. 2020.
\newblock Instance Adaptive Self-Training for Unsupervised Domain Adaptation.
\newblock In \emph{Proceedings of the European Conference on Computer Vision
  (ECCV)}, 415--430.

\bibitem[{Minaee et~al.(2021)Minaee, Boykov, Porikli, Plaza, Kehtarnavaz, and
  Terzopoulos}]{minaee2021image}
Minaee, S.; Boykov, Y.~Y.; Porikli, F.; Plaza, A.~J.; Kehtarnavaz, N.; and
  Terzopoulos, D. 2021.
\newblock Image Segmentation Using Deep Learning: A Survey.
\newblock \emph{IEEE Transactions on Pattern Analysis and Machine
  Intelligence}.

\bibitem[{Pan et~al.(2020)Pan, Shin, Rameau, Lee, and Kweon}]{intrada}
Pan, F.; Shin, I.; Rameau, F.; Lee, S.; and Kweon, I.~S. 2020.
\newblock Unsupervised Intra-domain Adaptation for Semantic Segmentation
  through Self-Supervision.
\newblock In \emph{Proceedings of the IEEE/CVF Conference on Computer Vision
  and Pattern Recognition (CVPR)}, 3763--3772.

\bibitem[{Richter et~al.(2016)Richter, Vineet, Roth, and Koltun}]{gta5}
Richter, S.~R.; Vineet, V.; Roth, S.; and Koltun, V. 2016.
\newblock Playing for Data: Ground Truth from Computer Games.
\newblock In \emph{Proceedings of the European Conference on Computer Vision
  (ECCV)}, 102--118.

\bibitem[{Ronneberger, Fischer, and Brox(2015)}]{unet}
Ronneberger, O.; Fischer, P.; and Brox, T. 2015.
\newblock U-Net: Convolutional Networks for Biomedical Image Segmentation.
\newblock In \emph{Proceedings of the International Conference on Medical Image
  Computing and Computer Assisted Intervention}, 234--241.

\bibitem[{Ros et~al.(2016)Ros, Sellart, Materzynska, Vazquez, and
  Lopez}]{synthia}
Ros, G.; Sellart, L.; Materzynska, J.; Vazquez, D.; and Lopez, A.~M. 2016.
\newblock The SYNTHIA Dataset: A Large Collection of Synthetic Images for
  Semantic Segmentation of Urban Scenes.
\newblock In \emph{Proceedings of the IEEE/CVF Conference on Computer Vision
  and Pattern Recognition (CVPR)}, 3234--3243.

\bibitem[{Saito et~al.(2018)Saito, Watanabe, Ushiku, and
  Harada}]{Saito_2018_CVPR}
Saito, K.; Watanabe, K.; Ushiku, Y.; and Harada, T. 2018.
\newblock Maximum Classifier Discrepancy for Unsupervised Domain Adaptation.
\newblock In \emph{Proceedings of the IEEE/CVF Conference on Computer Vision
  and Pattern Recognition (CVPR)}, 3723--3732.

\bibitem[{Saporta et~al.(2020)Saporta, Vu, Cord, and
  P{\'e}rez}]{saporta2020esl}
Saporta, A.; Vu, T.-H.; Cord, M.; and P{\'e}rez, P. 2020.
\newblock ESL: Entropy-guided Self-supervised Learning for Domain Adaptation in
  Semantic Segmentation.
\newblock In \emph{Proceedings of the IEEE/CVF Conference on Computer Vision
  and Pattern Recognition (CVPR) Workshops}.

\bibitem[{Simonyan and Zisserman(2015)}]{vgg}
Simonyan, K.; and Zisserman, A. 2015.
\newblock Very Deep Convolutional Networks for Large-Scale Image Recognition.
\newblock In \emph{Proceedings of the International Conference on Learning
  Representations}.

\bibitem[{Sun and Saenko(2016)}]{sun2016deep}
Sun, B.; and Saenko, K. 2016.
\newblock Deep CORAL: Correlation Alignment for Deep Domain Adaptation.
\newblock In \emph{Proceedings of the European Conference on Computer Vision
  (ECCV)}, 443--450.

\bibitem[{Tsai et~al.(2018)Tsai, Hung, Schulter, Sohn, Yang, and
  Chandraker}]{adaptseg}
Tsai, Y.-H.; Hung, W.-C.; Schulter, S.; Sohn, K.; Yang, M.-H.; and Chandraker,
  M. 2018.
\newblock Learning to Adapt Structured Output Space for Semantic Segmentation.
\newblock In \emph{Proceedings of the IEEE/CVF Conference on Computer Vision
  and Pattern Recognition (CVPR)}, 7472--7481.

\bibitem[{Vu et~al.(2019{\natexlab{a}})Vu, Jain, Bucher, Cord, and
  P{\'e}rez}]{advent}
Vu, T.-H.; Jain, H.; Bucher, M.; Cord, M.; and P{\'e}rez, P.
  2019{\natexlab{a}}.
\newblock ADVENT: Adversarial Entropy Minimization for Domain Adaptation in
  Semantic Segmentation.
\newblock In \emph{Proceedings of the IEEE/CVF Conference on Computer Vision
  and Pattern Recognition (CVPR)}, 2517--2526.

\bibitem[{Vu et~al.(2019{\natexlab{b}})Vu, Jain, Bucher, Cord, and
  P{\'e}rez}]{dada}
Vu, T.-H.; Jain, H.; Bucher, M.; Cord, M.; and P{\'e}rez, P.
  2019{\natexlab{b}}.
\newblock DADA: Depth-aware Domain Adaptation in Semantic Segmentation.
\newblock In \emph{Proceedings of the IEEE/CVF International Conference on
  Computer Vision (ICCV)}, 7364--7373.

\bibitem[{Wang et~al.(2020)Wang, Shen, Zhang, Duan, and Mei}]{fada}
Wang, H.; Shen, T.; Zhang, W.; Duan, L.-Y.; and Mei, T. 2020.
\newblock Classes Matter: A Fine-grained Adversarial Approach to Cross-domain
  Semantic Segmentation.
\newblock In \emph{Proceedings of the European Conference on Computer Vision
  (ECCV)}, 642--659.

\bibitem[{Wrenninge and Unger(2018)}]{synscapes}
Wrenninge, M.; and Unger, J. 2018.
\newblock Synscapes: A Photorealistic Synthetic Dataset for Street Scene
  Parsing.
\newblock \emph{arXiv:1810.08705}.

\bibitem[{Xiao et~al.(2018)Xiao, Liu, Zhou, Jiang, and Sun}]{upernet}
Xiao, T.; Liu, Y.; Zhou, B.; Jiang, Y.; and Sun, J. 2018.
\newblock Unified Perceptual Parsing for Scene Understanding.
\newblock In \emph{Proceedings of the European Conference on Computer Vision
  (ECCV)}, 418--434.

\bibitem[{Zhang et~al.(2018)Zhang, Dana, Shi, Zhang, Wang, Tyagi, and
  Agrawal}]{encnet}
Zhang, H.; Dana, K.; Shi, J.; Zhang, Z.; Wang, X.; Tyagi, A.; and Agrawal, A.
  2018.
\newblock Context Encoding for Semantic Segmentation.
\newblock In \emph{Proceedings of the IEEE/CVF Conference on Computer Vision
  and Pattern Recognition (CVPR)}, 7151--7160.

\bibitem[{Zhang et~al.(2021)Zhang, Zhang, Zhang, Chen, Wang, and Wen}]{proda}
Zhang, P.; Zhang, B.; Zhang, T.; Chen, D.; Wang, Y.; and Wen, F. 2021.
\newblock Prototypical pseudo label denoising and target structure learning for
  domain adaptive semantic segmentation.
\newblock In \emph{Proceedings of the IEEE/CVF Conference on Computer Vision
  and Pattern Recognition}, 12414--12424.

\bibitem[{Zhang et~al.(2019)Zhang, Zhang, Liu, and Tao}]{zhang2019category}
Zhang, Q.; Zhang, J.; Liu, W.; and Tao, D. 2019.
\newblock Category Anchor-Guided Unsupervised Domain Adaptation for Semantic
  Segmentation.
\newblock In \emph{Proceedings of the Conference on Neural Information
  Processing Systems}, 433--443.

\bibitem[{Zhao et~al.(2017)Zhao, Shi, Qi, Wang, and Jia}]{pspnet}
Zhao, H.; Shi, J.; Qi, X.; Wang, X.; and Jia, J. 2017.
\newblock Pyramid Scene Parsing Network.
\newblock In \emph{Proceedings of the IEEE/CVF Conference on Computer Vision
  and Pattern Recognition (CVPR)}, 2881--2890.

\bibitem[{Zheng and Yang(2020)}]{zheng2020unsupervised}
Zheng, Z.; and Yang, Y. 2020.
\newblock Rectifying Pseudo Label Learning via Uncertainty Estimation for
  Domain Adaptive Semantic Segmentation.
\newblock \emph{International Journal of Computer Vision (IJCV)}.

\bibitem[{Zou et~al.(2018)Zou, Yu, Kumar, and Wang}]{zou2018unsupervised}
Zou, Y.; Yu, Z.; Kumar, B.; and Wang, J. 2018.
\newblock Unsupervised Domain Adaptation for Semantic Segmentation via
  Class-Balanced Self-Training.
\newblock In \emph{Proceedings of the European Conference on Computer Vision
  (ECCV)}, 289--305.

\bibitem[{Zou et~al.(2019)Zou, Yu, Liu, Kumar, and Wang}]{zou2019confidence}
Zou, Y.; Yu, Z.; Liu, X.; Kumar, B.; and Wang, J. 2019.
\newblock Confidence Regularized Self-Training.
\newblock In \emph{Proceedings of the IEEE/CVF International Conference on
  Computer Vision (ICCV)}, 5982--5991.

\end{thebibliography}

\end{document}